\documentclass[letterpaper, 10 pt, conference]{ieeeconf}  

\IEEEoverridecommandlockouts                              

\usepackage{graphics} 
\usepackage{epsfig} 
\usepackage{mathptmx} 
\usepackage{times} 
\usepackage{amsmath} 
\usepackage{amssymb}  

\usepackage{bbm}
\usepackage{multirow}
\usepackage{booktabs}
\usepackage{hyperref}
\usepackage[capitalise]{cleveref}
\usepackage[table]{xcolor}

\title{\LARGE \bf
Enhancing Online Road Network Perception and Reasoning with Standard Definition Maps
}

\author{Hengyuan Zhang$^{*1}$, David Paz$^{*2}$, Yuliang Guo$^{2}$, Arun Das$^{2}$, Xinyu Huang$^{2}$, \\Karsten Haug$^{3}$, Henrik I. Christensen$^{1}$ and Liu Ren$^{2}$
\thanks{* These authors contribute equally to this work.}
\thanks{$^{1}$Hengyuan Zhang and Henrik I. Christensen are with the Contextual Robotics Institute, UC San Diego, 9500 Gilman Drive, La Jolla, CA 92122, USA
        {\tt\small \{hyzhang, hichristensen\}@ucsd.edu}.}%
\thanks{$^{2}$David Paz, Yuliang Guo, Arun Das, Xinyu Huang and Liu Ren are with Bosch North America and Bosch Center for AI (BCAI), 384 Santa Trinita Ave, Sunnyvale, CA 94085, USA
        {\tt\small \{david.pazruiz, yuliang.guo2, arun.das, xinyu.huang, liu.ren\}@us.bosch.com}.}%
\thanks{$^{3}$Karsten Haug is with Robert Bosch GmbH, Hessbruehlstrasse 21, Stuttgart-Baihingen Bade-Wuerttemberg 70565, Germany
        {\tt\small karsten.haug@de.bosch.com}.}%
\thanks{\copyright 2024 IEEE.  Personal use of this material is permitted.  Permission from IEEE must be obtained for all other uses, in any current or future media, including reprinting/republishing this material for advertising or promotional purposes, creating new collective works, for resale or redistribution to servers or lists, or reuse of any copyrighted component of this work in other works.}%
}

\begin{document}

\maketitle
\thispagestyle{empty}
\pagestyle{empty}


\begin{abstract}
Autonomous driving for urban and highway driving applications often requires High Definition (HD) maps to generate a navigation plan. Nevertheless, various challenges arise when generating and maintaining HD maps at scale. While recent online mapping methods have started to emerge, their performance especially for longer ranges is limited by heavy occlusion in dynamic environments. With these considerations in mind, our work focuses on leveraging lightweight and scalable priors--Standard Definition (SD) maps--in the development of online vectorized HD map representations. We first examine the integration of prototypical rasterized SD map representations into various online mapping architectures. Furthermore, to identify lightweight strategies, we extend the OpenLane-V2 dataset with OpenStreetMaps and evaluate the benefits of graphical SD map representations. A key finding from designing SD map integration components is that SD map encoders are model agnostic and can be quickly adapted to new architectures that utilize bird's eye view (BEV) encoders. Our results show that making use of SD maps as priors for the online mapping task can significantly speed up convergence and boost the performance of the online centerline perception task by 30\% (mAP). Furthermore, we show that the introduction of the SD maps leads to a reduction of the number of parameters in the perception and reasoning task by leveraging SD map graphs while improving the overall performance. Project Page: \href{https://henryzhangzhy.github.io/sdhdmap/}{https://henryzhangzhy.github.io/sdhdmap/}.
\end{abstract}    
\section{Introduction}
\label{sec:intro}

\begin{figure}
\centering
\includegraphics[width=1.0\linewidth]{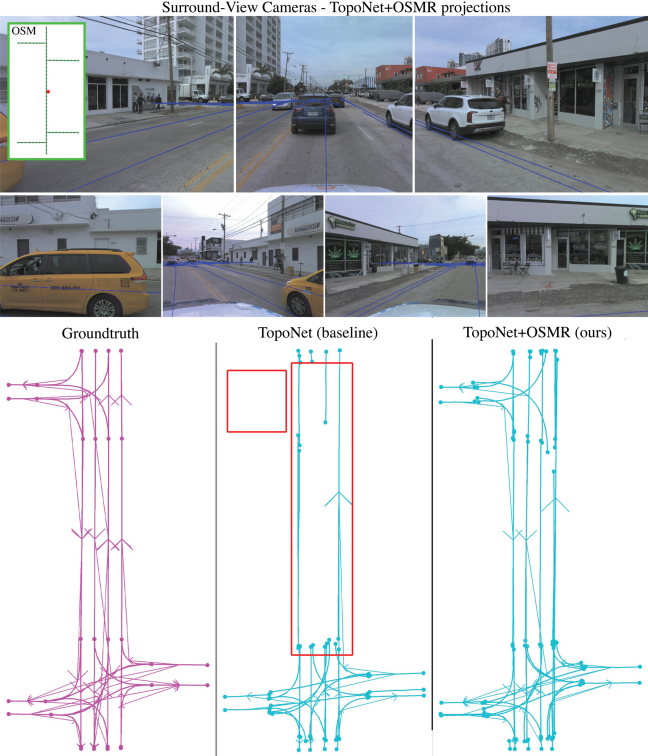}
\caption{Online road network perception and reasoning is challenging due to occlusion by on-road objects, especially at long-range as required by planning. In this example, the left turn map elements are heavily occluded by the vehicles. The baseline (TopoNet) using only image data misses the left turn while our method (TopoNet+OSMR--leveraging rasterized Standard Definition (SD) maps as the prior) predicts it correctly. Visualizations represent centerlines with connectivity information.}
\label{fig:topo-osmr-visuals}
\end{figure}

Research and development in the areas of autonomous driving has rapidly evolved over the past few decades.
Nevertheless, highly detailed and rich maps, often referred to as High Definition (HD) maps, have become a key component required by most perception and planning modules, such as in Autoware~\cite{Darweesh2021OpenPlanner2} and Apollo~\cite{fan2018baidu}. It is especially critical for the fully autonomous driving scenarios, when online perception of individual lanes~\cite{guo2020gen,9460822} cannot satisfy the need of complex motion planning or long-term navigation. Today, HD maps serve as the de facto modality and provide static contextual information for behavior prediction~\cite{gao2020vectornet,SalzmannIvanovicEtAl2020,chai2020multipath} and road topology and connectivity information for route planning and motion planning~\cite{zeng2019end,sadat2020perceive}. 

However, HD maps present significant challenges in terms of cost, scalability, and maintenance~\cite{Jiao18MLHDMap}. HD maps often require dedicated mapping teams to gather, process, and label data for the regions within the operational design domain. If a significant change occurs that displaces the original definitions, the original map must be updated which presents high maintenance overhead and potential failure points in the software. This overall process often requires human supervision, manual labeling, and extensive verification. As a result, these considerations present a bottleneck in terms of providing a scalable and cost-efficient solution. 

Therefore, cost-effective and online methods present benefits that can potentially address the pain points of HD maps. More recently a research effort has focused on the online component~\cite{li2021hdmapnet, liu2023vectormapnet, MapTR, maptrv2}, which appears to advance the classic lane detection by recovering the topological outputs as close to the HD maps as possible. For instance, MapTR~\cite{MapTR} introduces a method for online vectorized HD map generation; this work focuses on the perception of road boundaries, pedestrian crossings, and lane marks. As an extension to the perception task, Li et al.~\cite{li2023toponet} introduce a method termed TopoNet to model the underlying topology and road network connectivity for urban driving tasks. This approach focuses on the perception and reasoning task jointly and aims to reduce the gap between the features generated by fully online models and HD maps. Nevertheless, these methods present high computing requirements and still experience challenges in highly dynamic and occluded environments, especially for longer ranges which is required for planning.

To further explore online HD mapping for autonomous driving applications, our work seeks to incorporate lightweight priors as part of the formulation to improve performance while reducing computing complexity. More specifically, we explore the benefits of utilizing coarse priors in the form of the widely available Standard Definition (SD) maps, such as Google Maps and OpenStreetMaps (OSM)~\cite{haklay2008openstreetmap}. We evaluate the performance and computing implications of SD map representations and encoders for the detection and reasoning tasks. We perform our experiments using various architectures across open-source datasets and additionally introduce a new SD map dataset based on OSM to explore the benefits of graph-based SD map representations. In summary, our key contributions are as follows.
\begin{itemize}
\item We introduce different types of lightweight SD map representations into the online mapping tasks. We show that SD maps provide long-range prior information and can visually improve occluded regions; thus, resulting in better overall quantitative results. 

\item We investigate prototypical representations of the SD maps and the most effective integration with online map baselines with various architectures. 

\item Additionally, we expand the OpenLane-V2 dataset with OSM data to enable using SD maps as graph representations for graph-based architectures. Our dataset, termed OpenLane-V2-OSM, will be public. 

\end{itemize}

\section{Related Work}
\label{sec:related-work}

\begin{figure*}[t]
  \centering
   \includegraphics[width=.85\textwidth]{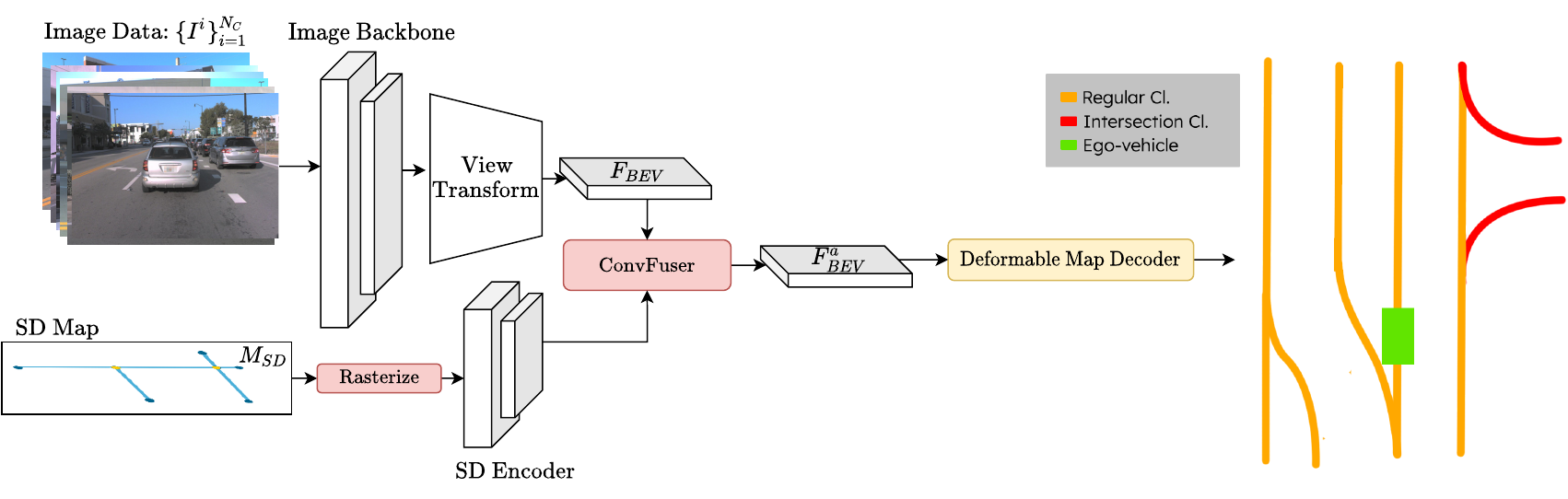}
   \caption{Our pipeline integrating a rasterized SD map with the state-of-the-art online perception approach MapTR. The model encodes the SD Map in rasterized features in bird's-eye view (BEV) space, and fuse them with image BEV features, and predicts centerlines with a deformable attention decoder.}
   \label{fig:perception-approach}
\end{figure*}

\textbf{High Definition (HD) Maps.} HD maps are costly to generate and require constant maintenance~\cite{Jiao18MLHDMap, zhang2023probabilistic}. To tackle these problems, two major directions emerge. One direction, HD map automation~\cite{9697426, 10073954, zhou2021automatic}, investigates automated HD map generation and map change detection and merging with a fleet. The other direction involves online HD mapping~\cite{li2021hdmapnet, liu2023vectormapnet, MapTR, maptrv2, li2023toponet}, which entails building an HD map on the fly. 

For HD map automation, the aspect more related to our work is online map generation. Early work focuses on drivable area or lane semantics~\cite{zhang2023probabilistic} that are important for navigation. Related methods introduce BEV-level scene understanding strategies using monocular camera data~\cite{9697426,10073954}.
Zhou et al.~\cite{zhou2021automatic} use instance segmentation for lane segmentation and a particle filter to extract lane information. The final vector map is then generated with the information from the OpenStreetMap (OSM)~\cite{haklay2008openstreetmap}.

HD map automation still requires a large fleet to constantly maintain them. To address this issue, online HD mapping has become more popular in recent years. STSU~\cite{can2021structured} builds BEV centerline road networks from a single camera. HDMapNet~\cite{li2021hdmapnet} takes surrounding view images and first generates semantic segmentation, then post-processes them to construct vectorized maps. This post-processing step is removed in VectorMapNet~\cite{liu2023vectormapnet} by directly generating vectorized representations using a transformer decoder. Furthermore, MapTR~\cite{MapTR} and MapTRv2~\cite{maptrv2} propose a permutation invariant loss for better learning map elements that are not directional. These works typically focus on the $[-30m, 30m]$ range and the performance deteriorates significantly when the range increases.

To make HD maps more useful for downstream tasks such as prediction and planning, they need more than simple map elements. The list extends to traffic elements such as traffic lights/signs and a stronger association component that reasons about their relationship. Building on top of prior large datasets with HD maps such as Nuscenes~\cite{caesar_nuscenes_2020} and Argoverse~\cite{Argoverse2}, OpenLane-V2~\cite{wang2023openlanev2} provides additional annotations including traffic light colors, turn signals, and their association to specific lanes. TopoNet~\cite{li2023toponet} uses this dataset to decode centerlines from BEV features, and applies a Scene Graph Neural Network (SGNN) to learn the final centerline and control relationships. \\

\textbf{Standard Definition (SD) Maps.} SD maps such as Google Maps and Open Street Maps include high-level road network information without lane-level information. They are scalable and lightweight solutions widely used in human driving that provide context for driving; these qualities can help address some of the issues in real-time perception such as occlusion. However, few methods use SD maps as prior for detailed mapping. In~\cite{zhou2021automatic}, they infer the connectivity of estimated lanes in intersections based on the connectivity of OSM. OSM are also used in~\cite{Hecker_2018_ECCV_e2e_osm} and~\cite{Amini2019e2eosm} towards e2e autonomous driving. Various works also use SD maps as context for downstream tasks such as prediction~\cite{liao2023osm} and planning~\cite{paz2023osm}. Inspired by these strategies, we explored using OSM as context for online HD mapping.
\section{Methodology}
\label{sec:sd-maps}

To evaluate the effectiveness of introducing SD maps as a prior for online HD mapping, we integrate SD maps into recent online mapping tasks. These tasks can be divided into two folds. One is the perception task, focusing on the map element such as lane line, road boundary, crosswalk and centerline prediction~\cite{MapTR,maptrv2}. The other adds reasoning to perception, which not only detects the map elements but also traffic elements such as traffic lights/signs and their relations~\cite{li2023toponet}. We describe our approaches to integrating SD maps into state-of-the-art models for the pure perception task in~\cref{sec:perception-task} and the joint perception and reasoning tasks in~\cref{sec:perception-reasoning-task}.

\subsection{Perception Task} \label{sec:perception-task}

For the perception task, we predict centerlines based on surround-view images and SD maps. More formally, given a sequence of image inputs $\{I^i\}_{i=1}^{N_C}$ generated from $N_C$ surround-view cameras, and an ego-centric SD map representation $M_{SD}$, the task predicts the centerlines $D_C$. A centerline predicted $D_C^k \in D_C$ is represented as a 2D line $\{(x_j^k,y_j^k)\}_{j=1}^{N_L}$ with $N_L$ waypoints in the ego-vehicle frame. Additionally, each centerline prediction contains an attribute to denote if a centerline is an intersection segment or a regular segment.

\textbf{Architecture.} We incorporate SD maps into the state-of-the-art online HD mapping architecture from MapTR~\cite{MapTR} which is based on an encoder-decoder architecture (\cref{fig:perception-approach}). An image encoder encodes surround-view images $\{I^i\}_{i=1}^{N_C}$ into perspective-view (PV) features $F_{PV}$. These PV features are then transformed into unified BEV features $F_{BEV}$ by a BEV view transform module. The BEV features are further decoded into map elements $D_C$. 

\textbf{Rasterized SD Map.} Given that SD maps are naturally represented in BEV, we fuse SD map features with BEV image features. We rasterize the SD map and generate a BEV SD map $M_{SD}$, each SD map class represented by a distinct color. An SD encoder, in this case a ResNet-18~\cite{he2016deep}, is employed to extract SD map features $F_{SD}$. We chose a lightweight encoder given that the features are already color-coded by semantics. 

SD map feature $F_{SD}$ is then interpolated and concatenated with the BEV feature from images $F_{BEV}$ along the channel dimension. Our design leverages this approach to align spatial features from BEV and SD maps together. The intuition revolves around using the SD map canvas to reduce the centerline search space in BEV. Subsequently, similar to fusing with BEV LiDAR feature in~\cite{liu2023vectormapnet, MapTR}, the ConvFuser with a simple two-layer convolutional neural network fuses the concatenated feature and output the fused BEV feature $F^a_{BEV}$. 

\textbf{Losses.} The losses are the same as introduced in~\cite{MapTR}, a combination of classification loss $\mathcal{L}_{cls}$, point distance loss $\mathcal{L}_{p2p}$ and edge directional loss $\mathcal{L}_{dir}$.

\subsection{Perception and Reasoning Task}
\label{sec:perception-reasoning-task}

\begin{figure*}[t]
  \vspace{0.14cm}
  \centering
   \includegraphics[width=.9\textwidth]{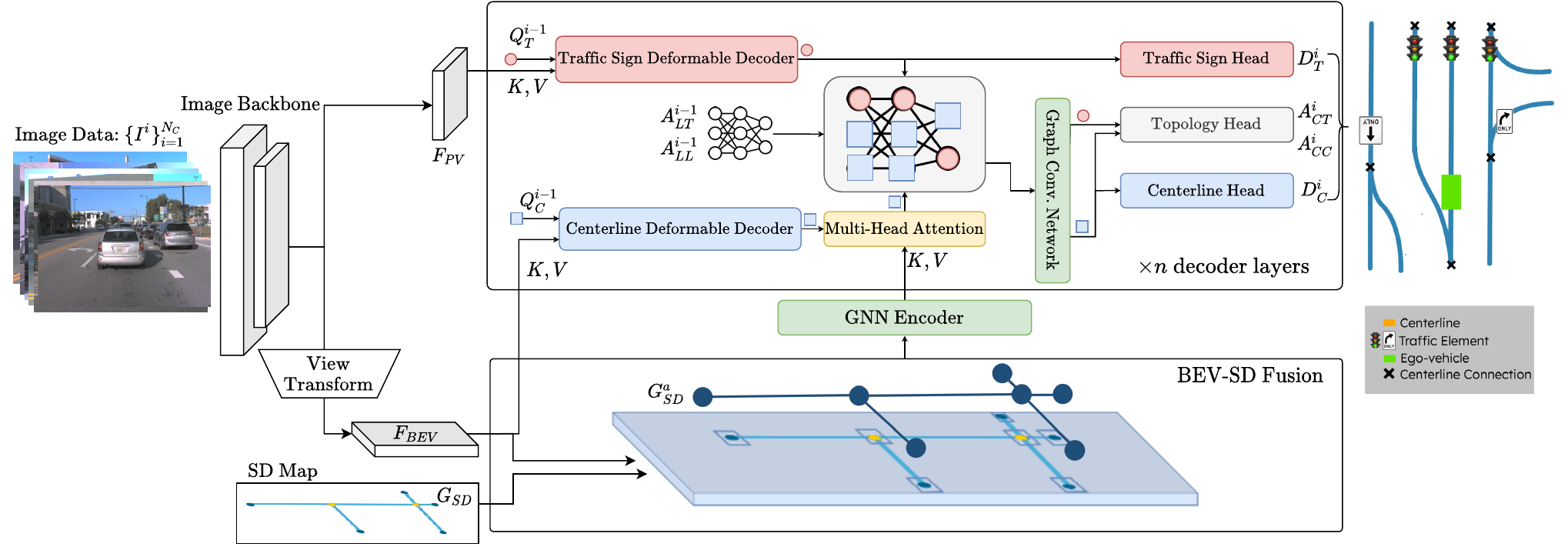}
   \caption{Our pipeline integrating graph-based SD maps with the state-of-the-art perception and reasoning architecture based on TopoNet with a BEV-SD OSM graph encoder. The method processes multi-view image data, OSM SD map graphs, and leverages deformable decoders along with a Scene Graph Neural Network process to predict centerlines, traffic elements, and their relationships.}
   \label{fig:perception-reasoning-approach}
\end{figure*}

This section builds on the task explored in~\cref{sec:perception-task} by incorporating a reasoning component to road network perception. In addition to identifying the centerline elements nearby, with respect to the ego-agent, this task seeks to identify the traffic elements on the road, the relationships between centerlines, and the relationships between the traffic elements and centerlines detected. Traffic elements, such as traffic lights, road markings, and road signs, provide important navigation information to HD maps.

More formally, given the same inputs $\{I^i\}_{i=1}^{N_C}$ and $M_{SD}$,
the task involves predicting the centerlines $D_C$, the road-side traffic elements $D_T$, and the relational attributes, the association matrix between centerlines and centerlines $A_{CC}$, and the association matrix between centerlines and traffic elements $A_{CT}$~\cite{wang2023openlanev2}. From this these outputs, a centerline predicted $D_C^i \in D_C$ is represented as a 3D line $\{(x^i,y^i,z^i)\}_{i=1}^{N_L}$ with $N_L$ waypoints in the ego-vehicle frame. A traffic element $j$ detected in image $i$ ($D_T^{i,j}$) can be represented as a 2D bounding box $(x, y, w, h, class)$, where $x, y$ denote position, $w, h$ bounding box dimensions, and $class$ denotes the traffic sign attribute such as $turn\_left$, $turn\_right$, $red, green, yellow$ for traffic lights. 


\textbf{Architecture.} We employ the TopoNet Architecture~\cite{li2023toponet} as the basis of our approach. The approach outlined in~\cref{fig:perception-reasoning-approach} utilizes an image backbone (i.e. ResNet-50) to process $N_C$ image inputs and generate their corresponding images features $F_{PV}$. These perspective view features are then utilized in multiple downstream components. First, a deformable attention decoder~\cite{zhu2020deformable} is used as a traffic element decoder where the traffic element queries $Q_{T}$ attend to the perspective view features $F_{PV}$ to decode traffic elements embeddings. Similarly, the image features $F_{PV}$ are processed by a BEVFormer Encoder~\cite{li2022bevformer} to transform the perspective view features into BEV features--this is denoted by $F_{BEV}$. In the following centerline deformable attention decoder, the centerline queries denoted by $Q_{C}$ then attend to the BEV features to generate centerline embeddings. 

A key difference with respect to TopoNet involves the added SD map feature encoder. In this section, we experiment with two encoders: one that processes SD maps as rasters and one that operates directly on SD map graphs. The raster-based encoder resizes the input SD map $M_{SD}$ to the same dimensions as the BEV feature map $F_{BEV}$ and stacks them together along the channel dimension; we utilize the encoder setup and rasterization process as introduced in~\cref{sec:perception-task}. In contrast, the graph-based encoder fuses SD map graphs with BEV features and is combined with the outputs from the centerline deformable decoder by leveraging a multi-head attention mechanism~\cite{vaswani2017attention}. In the following parts of this section, we discuss the encoder and alignment process for SD map graphs, the perception heads, and reasoning process between centerlines and traffic signs. The section concludes with the losses used in the training process.

\textbf{BEV-SD Graph Fusion.}\label{bev-sd-fusion} The BEV-SD fusion component shown in~\cref{fig:perception-reasoning-approach} leverages $F_{BEV}$ to augment the node features from a given SD map graph $G_{SD}=(V_{SD}, E_{SD})$, where $V_{SD}=\{1,...,n\}$ and $E_{SD} \subseteq V_{SD} \times V_{SD}$. More specifically, vertex $i$ corresponds to node feature $X^i_{SD}$ which contains positional information with respect to the ego-vehicle, namely $X^i_{SD}=(x_i, y_i)$. Since $F_{BEV} \in \mathbb{R}^{H_B \times W_B \times C_B}$ also encodes spatial BEV features in an ego-centric perspective with a fixed perception range given by an $H_B \times W_B$ grid, we can align a given SD node $X^i_{SD}$ in BEV by scaling $(x_i, y_i)$ as shown in~\cref{eq3.2.1} and~\cref{eq3.2.2}, where $H^m_B$ and $W^m_B$ are convertion factors in terms of $cell/m$. Thus, the BEV feature corresponding to a given SD map element located at $(x_i, y_i)$ can be indexed by $(x^B_i, y^B_i)$. Similar to the rasterized encoder counterpart, our design is motivated by aiming to spatially align BEV and SD map features--which intuitively translates to using SD maps as a reference to regress centerlines and not start from scratch.

\begin{equation}
\label{eq3.2.1}
x^B_i = \lfloor x_i \cdot H^m_B \rfloor + \frac{H_B}{2}
\end{equation}

\begin{equation}
\label{eq3.2.2}
y^B_i = \lfloor y_i \cdot W^m_B \rfloor  + \frac{W_B}{2}
\end{equation}

This node-level feature augmentation process is performed for every SD map node within the BEV perception range by concatenating the initial node feature and the BEV feature corresponding that that position as shown in~\cref{eq3.2.3}. The augmented SD map graph is denoted by $G^a_{SD}$.

\begin{equation}
\label{eq3.2.3}
X^{a,i}_{SD} = \operatorname{concat}\left(X^i_{SD}, F_{BEV}(x_i^B,y_i^B)\right)
\end{equation}
\mbox{}

\textbf{SD Map Graph Encoder.} To process the augmented SD map graph $G^a_{SD}$, we employ an Edge Convolution graph encoder as it has been shown to effectively capture local geometric attributes~\cite{wang2019dynamic}. Our GNN formulation leverages the road connectivity information provided in the form of an adjacency matrix to determine node $i$'s neighbors.

This GNN approach then utilizes a two-layer Multi-layer Perceptron (MLP) with a ReLU activation function to extract relevant features from an augmented representation, $\left[X_{SD}^{a,i}, X_{SD}^{a,j}-X_{SD}^{a,i}\right]$. The features are subsequently averaged across all $\mathcal{N}(i)$ neighbors as shown in~\cref{eq3.2.4}.

\begin{equation}
\label{eq3.2.4}
X_{SD}^{a,i}=\frac{1}{\mathcal{N}(i)}\sum _{j \in \mathcal{N}(i)} \operatorname{MLP}_\theta\left(\left[X_{SD}^{a,i}, X_{SD}^{a,j}-X_{SD}^{a,i}\right]\right)
\end{equation}

After the graph propagation process, the SD map node features are attended by the output queries from the Centerline Deformable Decoder as denoted by the yellow block in~\cref{fig:perception-reasoning-approach}.

A Scene Graph Convolutional Neural Network (SGNN) then takes the output from previous stages to capture the relational attributes among centerlines and between centerlines and traffic elements, as introduced in TopoNet~\cite{li2023toponet}.

\textbf{Losses.}\label{sec:perception-reasoning-losses} We utilize the loss formulation from~\cite{li2023toponet} to supervise the outputs generated at every decoder layer. There are two key components to the loss based on a Bipartite matching process~\cite{carion2020detr}  which includes a detection component and a reasoning loss component. The detection component uses an IOU loss, an L1 loss for bounding box regression and the Focal loss~\cite{lin2017focal} for classification of traffic elements in perspective view. Similarly, for centerline detection we use the Focal loss and the L1 loss. Finally, the reasoning component uses the Focal loss in the process of classifying correct relational assignments.

\section{Experiments}
\label{sec:experiments}

We perform extensive experiments to validate the effectiveness of SD maps in online HD mapping. We introduce the datasets and metrics in~\cref{sec:datasets} and~\cref{sec:metrics}. We subsequently present the experiments and results for the perception task in~\cref{sec:perception-results} and for the perception and reasoning task in~\cref{sec:perception-reasoning-results}.

\subsection{Datasets}\label{sec:datasets}

\textbf{OpenLane-V2.}
Our perception experiments are based on the OpenLane-V2 (OLV2) dataset (subset-A)~\cite{wang2023openlanev2}, which is based on the Argoverse 2 dataset~\cite{Argoverse2}. They provide ego-centric SD maps along with seven surround-view camera images. SD maps from OpenLane-V2 include three classes, $road$, $crosswalk$ and $sidewalk$. Each class is presented as a set of polylines. From the labels, only the centerline labels are used. They are represented as a set of points and are resampled to a fixed number of points $N_L$ following MapTR~\cite{MapTR}. In our experiments for the perception task, $N_L=20$.\\

\begin{figure}[htbp]
    \centering
    \includegraphics[width=1.0\linewidth]{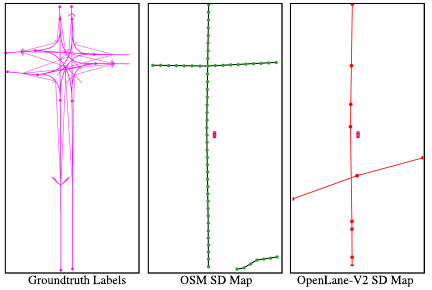}
    \caption{Visual comparison between the groundtruth online maps, OSM SD maps, and OpenLane-V2 (OLV2) SD maps. OSM SD maps appear to be more consistent with the groundtruth. }
    \label{fig:openlanev2-osm}
\end{figure}

\textbf{OpenLane-V2-OSM.}
In our experiments, we enhance the OLV2 dataset with OSM data—adding 1,000 maps from Argoverse 2 with WGS84 conversions. Each map includes full OSM attributes, and a post-processing step creates ego-centric SD graph representations for faster data loading. Both dataset representations will be publicly accessible. 

OSM offers lightweight yet diverse contextual information for driving scenarios. Although not adequate for direct lane-level navigation, it provides road-level details through \textit{way} and \textit{node} elements with various attributes. Node elements describe point-level features like stop signs, while way elements cover a range from small road segments to pedestrian crosswalks, including attributes like category types, speed limits, and lane numbers if applicable. A visualization is shown in~\cref{fig:openlanev2-osm}; where we observe groundtruth HD map labels, an OSM SD map, and an OLV2 SD map.

\subsection{Metrics}\label{sec:metrics}

For the perception-only task, we follow~\cite{MapTR} to evaluate the proposed architecture using Average Precision (AP) under Chamfer distance. The Chamfer distance gives the distance of two point sets as the average of the closest point distance. An association threshold determines whether a map element is considered a true positive. Three thresholds T$_1$=0.5m, T$_2$=1.0m, T$_3$=1.5m are used.

For the perception and reasoning task, we follow the metric from OLV2~\cite{wang2023openlanev2}. The OLV2 dataset uses the OLV2 score OLS, which is an average of the 3D lane detection score DET$_l$, the traffic element recognition score DET$_t$, the topology score between centerlines TOP$_{ll}$ and the topology score between centerlines and traffic elements TOP$_{lt}$.

\subsection{Perception Results} \label{sec:perception-results}

\begin{figure}
    \centering
    \includegraphics[width=\linewidth]{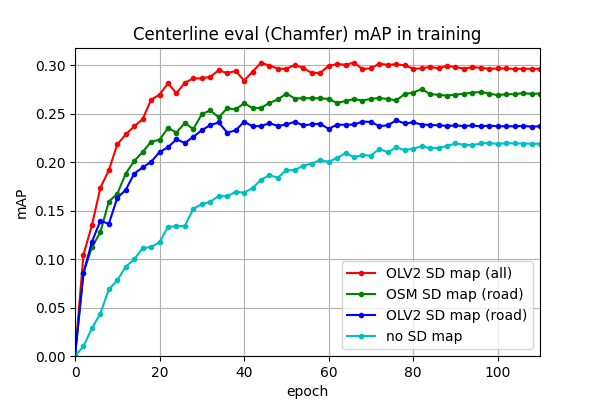}
    \caption{Evaluation mAP during training with Chamfer distance. The model with an SD map converges much faster and achieves better performance.}
    \label{fig:perception-task-sdmap-eval-mAP}
\end{figure}

We adapt MapTR to OLV2 dataset as the baseline. The perception range is increased from $[-30m, 30m]$ to $[-50m, 50m]$ range. The increase in perception range presents significant challenges for MapTR. For our approaches with rasterized SD maps, we experiment with three type of SD maps. OLV2 ($R$, $CW$, $SW$) has all three classes SD map features $road$, $crosswalk$ and $sidewalk$, OLV2 ($R$) only maintains $road$ features and OSM ($R$) has $road$ features extracted from OSM. These models predict regular and intersection centerlines as a two-class problem. During evaluation, we measure the collective average performance between the two classes since separate benchmarks result in a negligible difference. 

As shown in~\cref{fig:perception-task-sdmap-eval-mAP}, integrating with rasterized SD map with all classes makes the training converge 10x faster: at epoch 10, the method that uses SD maps reaches similar performance to the model without SD maps but trains for 110 epochs. OLV2 ($R$, $CW$, $SW$) also presents 30\% relative improvement in terms of mAP for the model trained for 110 epochs.

\begin{figure}[t]
  \vspace{0.14cm}
  \centering
   \includegraphics[width=1.\linewidth]{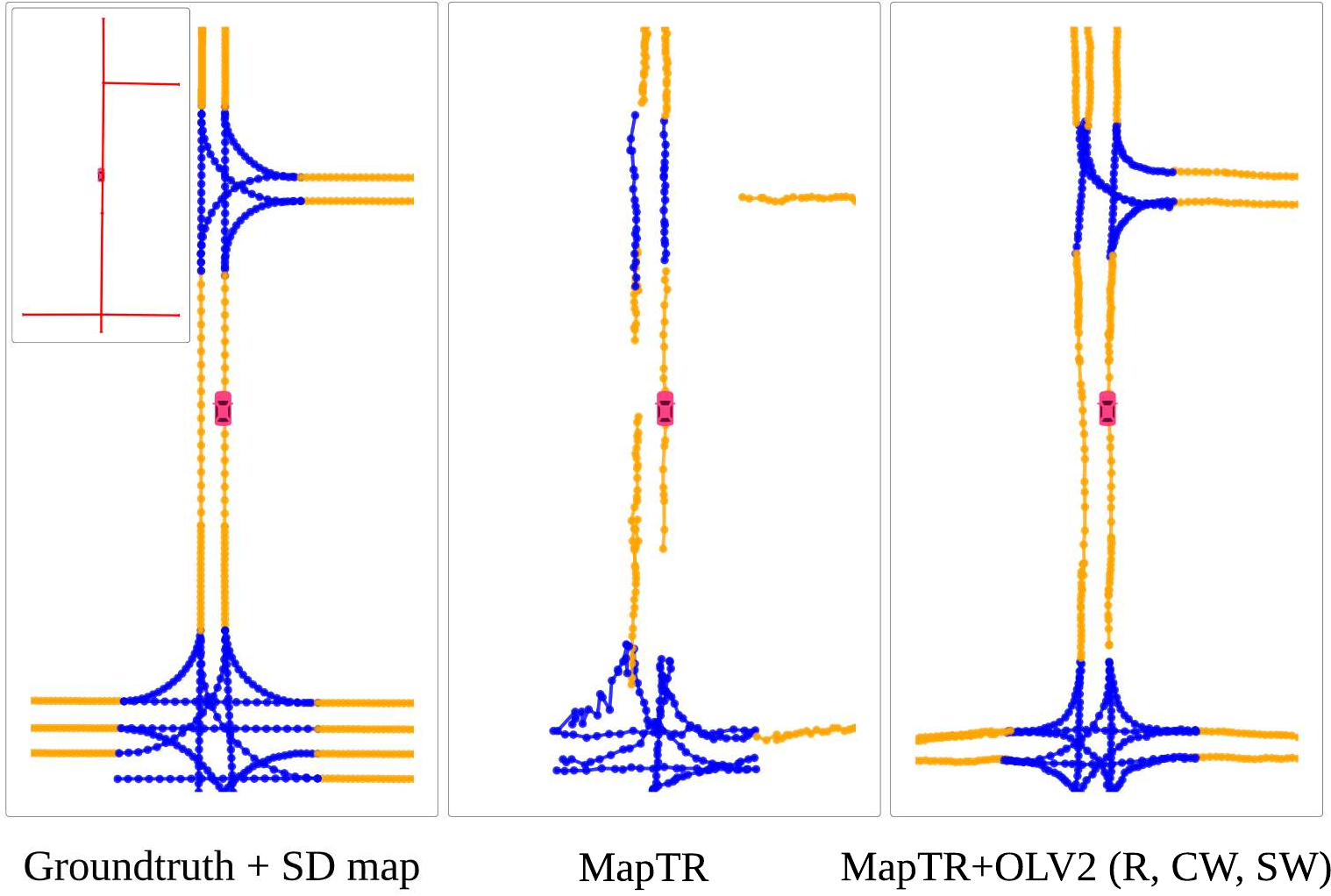}
   \caption{Qualitative comparison of perception-only methods with and without SD maps. Blue color for intersection or connectors, orange color otherwise. Both models are trained for 25 epochs.}
   \label{fig:perception-sdmap}
\end{figure}


As shown in~\cref{tab:perception-task-table}, among all three types of SD maps, we see consistent improvement from the baseline method. The OLV2 ($R$, $CW$, $SW$) with more SD map classes reaches the highest performance. For SD maps with only $road$ features, OSM prior gives better results than OLV2 prior. This can potentially be caused by misalignment of the SD maps and groundtruth. We observe this in both SD map types and an example is shown in~\cref{fig:openlanev2-osm}, where we observe missing SD map labels or undefined HD map features such as parking lot entrances that can cause inconsistencies.

In summary, these results suggest that SD maps provide useful priors for online HD mapping and speed up convergence and boost performance significantly. Qualitatively, we observe that our models with SD map priors perform better for far intersections, especially after the turning point where occlusion happens, as shown in~\cref{fig:perception-sdmap}. This is helpful for prediction and planning tasks to navigate intersections.

\begin{table}[t]
\centering
\scriptsize
\caption{Perception-only Task Results with Rasterized SD Maps
[Key: R = Road, CW = Crosswalk, SW = Sidewalk]
}
\resizebox{\columnwidth}{!}{%

\begin{tabular}{@{}llccccc@{}}
\toprule
& \multirow{2}{*}{SD Type}& \multirow{2}{*}{Epoch} & \multicolumn{4}{c}{Chamfer Distance AP} \\
 & &  & mAP  & T$_1$ & T$_2$ & T$_3$ \\
\midrule
MapTR & None    & \multirow{4}{*}{24} & 13.4 & 0.7  & 11.3 & 28.1 \\
MapTR+OLV2 & $R$   & & 22.3 & 4.4  & 23.2 & 39.4 \\
MapTR+OLV2 & $R$, $CW$, $SW$   & & \textbf{27.1} & \textbf{7.0}  & \textbf{28.8} & \textbf{45.5} \\
MapTR+OSMR & $OSM$ $R$  & & 23.0 & 5.1  & 24.2 & 39.8 \\
\midrule
MapTR & None    & \multirow{4}{*}{110} & 21.9 & 3.3  & 22.3 & 40.0 \\
MapTR+OLV2 & $R$   & & 23.7 & 6.2  & 24.5 & 40.4 \\
MapTR+OLV2 & $R$, $CW$, $SW$    & & \textbf{29.6} & \textbf{10.8}  & \textbf{31.4} & \textbf{46.6} \\
MapTR+OSMR & $OSM$ $R$  & & 27.1 & 9.4  & 28.6 & 43.2 \\
\bottomrule
\end{tabular}%
}
\label{tab:perception-task-table}
\end{table}

\subsection{Perception and Reasoning Results} \label{sec:perception-reasoning-results}

This section covers experiments conducted with rasterized and graph-based SD maps for the perception and reasoning task based on TopoNet~\cite{li2023toponet}. The experiments include performance trade-offs from rasterized vs graph-based methods as TopoNet facilitates integration of not just raster based representations but also vector/graph-based representations.

As shown in~\cref{tab:toponet-state-of-the-art}, leveraging SD map prior in either rasterized (OSMR) or graph representation (OSMG) leads to consistent improvements in the overall metrics. While other methods such as TopoMLP-YOLO~\cite{wu2023topomlp} and MFV-ViT-L~\cite{wu20231stplace_mfv} focus on improving results with better traffic element detectors and larger backbones, our experiments show that SD map can benefit lane perception and reasoning without significant changes to the architecture. We consider these approaches contributing in orthogonal directions compared to ours, thus do not compete with each other. However, we hypothesize that by utilizing SD maps as priors, a performance boost can be observed for TopoMLP, MFV, and other BEV-based architectures.

\begin{table}[h]
\centering
\scriptsize
\caption{Perception and Reasoning Results with SD Maps and comparisons with other methods.}
\resizebox{\columnwidth}{!}{%
\begin{tabular}{lcccccc}
\toprule
                & Backbone & OLS                    & DET$_l$           & DET$_t$           & TOP$_{ll}$        & TOP$_{lt}$\\ 
\midrule
TopoNet         & R50      & 34.8                   & 28.4              & 45.0              & 4.2               & 20.7 \\ 
\rowcolor{gray!20}TopoNet+OSMR    & R50      & \textbf{37.7}    & \textbf{30.6}     & 44.6              & \textbf{7.7}      & \textbf{22.9} \\ 
\rowcolor{gray!20}TopoNet+OSMG    & R50      & 36.7                   & 30.0              & \textbf{47.6}     & 5.4               & 21.3 \\ 
\midrule
TopoMLP         & R50 & 38.2 & 28.3 & 50.0 & 7.2 & 22.8 \\
TopoMLP-YOLO    & R50 & 41.2 & 28.8 & 53.3 & 7.8 & 30.1 \\
MFV-R50         & R50 & - & 18.2 & - & - & - \\
MFV-ViT-L       & ViT-L & \textbf{53.2} & \textbf{35.3} & \textbf{79.9} & \textbf{23.0} & \textbf{33.3} \\
\bottomrule
\end{tabular}%
}

\label{tab:toponet-state-of-the-art}
\end{table}

\textbf{Impact of rasterized SD map features.} To evaluate SD maps as rasters using the Toponet architecture, we replace the GNN SD map encoder introduced in~\cref{fig:perception-reasoning-approach} with the SD map encoder from the perception-only task covered in~\cref{sec:perception-task}. Furthermore, rather than utilizing the Multi-head Attention mechanism, we simply perform a feature augmentation to the BEV features $F_{BEV}$ along the channel dimension.

\begin{table}[h]
\centering
\scriptsize
\caption{Perception and Reasoning Results using Rasterized SD Maps as priors.
[Key: R = Road, CW = Crosswalk, SW = Sidewalk]
}
\resizebox{\columnwidth}{!}{%
\begin{tabular}{@{}lllllllll@{}}
\toprule
        & SD Type         & Param & OLS  & DET$_l$ & DET$_t$ & TOP$_{ll}$ & TOP$_{lt}$ & t (ms) \\ 
\midrule
TopoNet & None            & 62.9M & 34.8 & 28.4    & 45.0    & 4.15       & 20.7       & \textbf{388} \\ 
TopoNet+OLV2 & $R$             & 75.9M & 34.9 & 26.2    & 47.3    & 4.55       & 20.1      & 407 \\ 
TopoNet+OLV2 & $R$, $CW$, $SW$ & 75.9M & 36.1 & 27.9    & \textbf{48.1}    & 5.14       & 20.9      & 407 \\ 
TopoNet+OSMR & $OSM$ $R$      & 75.9M & \textbf{37.7} & \textbf{30.6} & 44.6 & \textbf{7.71} & \textbf{22.9} & 407 \\ \bottomrule
\end{tabular}%
}

\label{tab:toponet-rasters}
\end{table}

We perform an ablation with three different types of maps and features. Similar to~\cref{sec:perception-results}. 
We train each one of the models for 24 epochs and evaluate on the OLV2 benchmark. The results shown in~\cref{tab:toponet-rasters} provide an insight into the most relevant attribute types. For instance, even though we incorporate OLV2 SD maps in the first two experiments, the performance of centerline detection DET$_l$ decreases while the traffic element detection metric DET$_t$ indicates performance benefits. Since the OLV2 SD map road-level attributes are not available and the map generation process is not public, we hypothesize--as supported by~\cref{fig:openlanev2-osm}--that the performance gaps in centerline detection scores originate from the misalignment between groundtruth HD maps and the utilized SD maps. However, the performance benefits are observed in terms of traffic sign detection as crosswalks and sidewalks can provide contextual information on the location or existence of traffic elements. As an example, crosswalks are often located at intersections which are stop sign or traffic light protected road segments. Although the results may initially be counter intuitive given the experiments presented in~\cref{sec:perception-results} (\cref{tab:perception-task-table}), the TopoNet architecture (perception and reasoning) utilizes an SGNN component to capture relational attributes between centerline and traffic elements. These relationships are then taken into account when regressing centerlines and thus may be influencing their accuracy. We hypothesize that this underlying relationship between traffic elements and crosswalks/sidewalks may be helping boost performance in traffic element detection but also degrading centerline detection as centerlines are not specific to intersections only and can extend to regular road segments.

In contrast, the selected OpenStreetMap attributes are known and selected based on the urban driving operational design domain which aligns well with the OLV2 dataset. As a result, we observe improvements across all of the metrics introduced which include detection and reasoning scores. Lastly, we present qualitative results which portray the benefits of leveraging SD maps in occluded scenarios. In~\cref{fig:topo-osmr-visuals}, we present side-by-side comparisons of the TopoNet baseline model and our approach that makes use of rasterized OSM. We observe better alignment of centerline features despite of severe occlusion.

\textbf{Graph-based SD Maps.} We evaluate the performance of graph-based SD map encoders using various backbones in~\cref{tab:toponet-graphs}. The baseline model (TopoNet-R50) utilizes a ResNet-50~\cite{he2016deep} backbone. However, we additionally evaluate the performance using lighter backbones including ResNet-18 and ResNet-34. While Wu et al.~\cite{wu2023topomlp} show that recent backbones~\cite{lee2019energy,liu2021swin} can significantly increase the performance in the centerline detection task, our focus is on evaluating the trade-offs between performance and lighter weight methods such as SD map graphs.

In~\cref{tab:toponet-graphs}, the methods that utilize OSM SD map graphs are denoted by TopoNet-R\textbf{X}+OSMG, where \textbf{X} specifies different backbones. Evidently, as the number of parameters within a backbone increase, the performance across the different metrics also increases and is consistent across the corresponding TopoNet baselines. TopoNet-R34+OSMG is capable of boosting performance with respect to the larger Toponet-R50 while reducing the number of parameters and inference time. Furthermore, the larger TopoNet-R50+OSMG enables higher performance without significantly increasing the number of parameters or inference times. The overall results indicate that improvements can be achieved with lightweight methods with respect to the corresponding baselines. The inference times are measured on a Titan Xp GPU. 
The graph-based OSM encoder yields lower performance improvement than the rasterized OSM encoder but with significantly fewer parameters (+1.7M vs + 13M).

\begin{table}[h]
\vspace{0.12cm}
\centering
\scriptsize
\caption{Perception and Reasoning Results with OSM SD Map Graphs and Different Backbones}
\resizebox{\columnwidth}{!}{%

\begin{tabular}{@{}llllllll@{}}
\toprule
             & Param     & OLS           & DET$_l$       & DET$_t$       & TOP$_{ll}$    & TOP$_{lt}$    & t (ms) \\ 
\midrule
TopoNet-R18  & 49.8M     & 32.3          & 24.3          & 44.0          & 3.1           & 18.8          & \textbf{349}  \\
TopoNet-R18+OSMG & 51.6M     & 33.2          & 26.9          & 42.9          & 3.86          & 18.9          & 365    \\
TopoNet-R34  & 60.0M     & 33.3          & 26.2          & 43.8          & 3.63          & 19.6          & 361    \\
TopoNet-R34+OSMG & 61.7M     & 35.5          & 29.3          & 45.0          & 4.82          & 21.0          & 379   \\
TopoNet-R50  & 62.9M     & 34.8          & 28.4          & 45.0          & 4.15          & 20.7          & 388  \\
TopoNet-R50+OSMG & 64.6M     & \textbf{36.7} & \textbf{30.0} & \textbf{47.6} & \textbf{5.37} & \textbf{21.3} & 407 \\ 
\bottomrule
\end{tabular}%
}
\label{tab:toponet-graphs}
\end{table}




\textbf{Robustness to Localization Error.} To demonstrate robustness to localization errors we conduct an experiment by injecting SD map noise to translation and rotation. For translational error, we add fixed magnitude noise along random directions at the scale of 0.25m, 0.5m, 1.0m, and 2.0m (quite high for lanes), and for rotational error, we experiment with noise at the scale of 0, 5, 10 degrees. Our results in~\cref{table:localization} show that SD map localization errors with up to 1.0m in translation and 5 degrees in rotational error result in at most 5\% performance degradation. This demonstrates exceptional robustness to localization errors that can significantly affect lane-level perception.

\begin{table}[ht]
\centering
\scriptsize
\caption{Performance Loss with Various Localization Error}
\begin{tabular}{@{}l|rrr@{}}
\toprule
& \multicolumn{3}{c}{Rotational Error} \\
Translational Error   & 0 Degrees & 5 Degrees & 10 Degrees \\
\midrule
0.25m & 0.45\% & 2.33\% & 13.64\% \\
0.5m  & 0.90\% & 2.80\% & 14.18\% \\
1.0m  & 3.07\% & 5.01\% & 15.95\% \\
2.0m  & 10.82\% & 12.43\% & 21.14\% \\
\bottomrule
\end{tabular}

\label{table:localization}
\end{table}

\section{Conclusion}
\label{sec:conclusion}

We incorporate SD maps into online HD mapping for both perception-only task and joint perception and reasoning tasks. We show that SD maps can make centerline perception models converge significantly faster and achieve better performance. Adding them in graph form can reduce the model size while improving performance. We additionally curated and made public a dataset based on OSM. The effectiveness of the proposed methods, especially for longer ranges and occluded scenes, contributes to addressing the current online mapping challenges and scalability constraints from autonomous driving HD maps. To create better online maps, future research is needed to address inaccuracies in SD maps and to produce more consistent structural representations.

\addtolength{\textheight}{-0cm}   







\bibliographystyle{unsrt}
\bibliography{ref}

\end{document}